\documentclass[10pt,twocolumn,letterpaper]{article}

\pdfoutput=1

\usepackage{iccv}
\usepackage{times}
\usepackage{epsfig}
\usepackage{graphicx}
\usepackage{amsmath}
\usepackage{amssymb}

\let\origfootnote\footnote
\renewcommand{\footnote}[1]{\kern.06em\origfootnote{#1}}
\newcommand{\punctfootnote}[1]{\kern-.06em\origfootnote{#1}}

\usepackage{subcaption}
\usepackage{paralist}

\usepackage[pagebackref=true,breaklinks=true,letterpaper=true,colorlinks,bookmarks=false]{hyperref}

\DeclareGraphicsExtensions{.pdf,.jpg,.png}
\graphicspath{{figs/}}

\newcommand{\secref}[1]{Section~\ref{sec:#1}}
\newcommand{\figref}[1]{Figure~\ref{fig:#1}}

\iccvfinalcopy
\def\iccvPaperID{605} 

\ificcvfinal\fi
\begin{document}

\mathchardef\mhyphen="2D

\title{Wide-Area Image Geolocalization with Aerial Reference Imagery}

\author{
  \begin{minipage}{\linewidth}
    \centering
    \begin{minipage}{1.5in}
      \centering
      Scott Workman$^1$\\
      {\tt\small scott@cs.uky.edu}
    \end{minipage}
    \begin{minipage}{1.5in}
      \centering
      Richard Souvenir$^2$\\
      {\tt\small souvenir@uncc.edu}
    \end{minipage}
    \begin{minipage}{1.5in}
      \centering
      Nathan Jacobs$^1$\\
      {\tt\small jacobs@cs.uky.edu}
    \end{minipage}
    \\[.2cm]
    \begin{minipage}{1.8in}
      \centering
      $^1$University of Kentucky \\ 
    \end{minipage}
    \begin{minipage}{3in}
      \centering
      $^2$University of North Carolina at Charlotte \\
    \end{minipage}
  \end{minipage}
}

\maketitle

\begin{abstract}

We propose to use deep convolutional neural networks to address the
problem of cross-view image geolocalization, in which the geolocation
of a ground-level query image is estimated by matching to
georeferenced aerial images. We use state-of-the-art feature
representations for ground-level images and introduce a cross-view
training approach for learning a joint semantic feature representation for aerial
images. We also propose a network architecture that fuses features
extracted from aerial images at multiple spatial scales.  To support
training these networks, we introduce a massive database that contains
pairs of aerial and ground-level images from across the United States.
Our methods significantly out-perform the state of the art on two
benchmark datasets. We also show, qualitatively, that the proposed
feature representations are discriminative at both local and
continental spatial scales.

\end{abstract}

\section{Introduction}

\begin{figure}

  \centering

  \includegraphics[width=.95\linewidth]{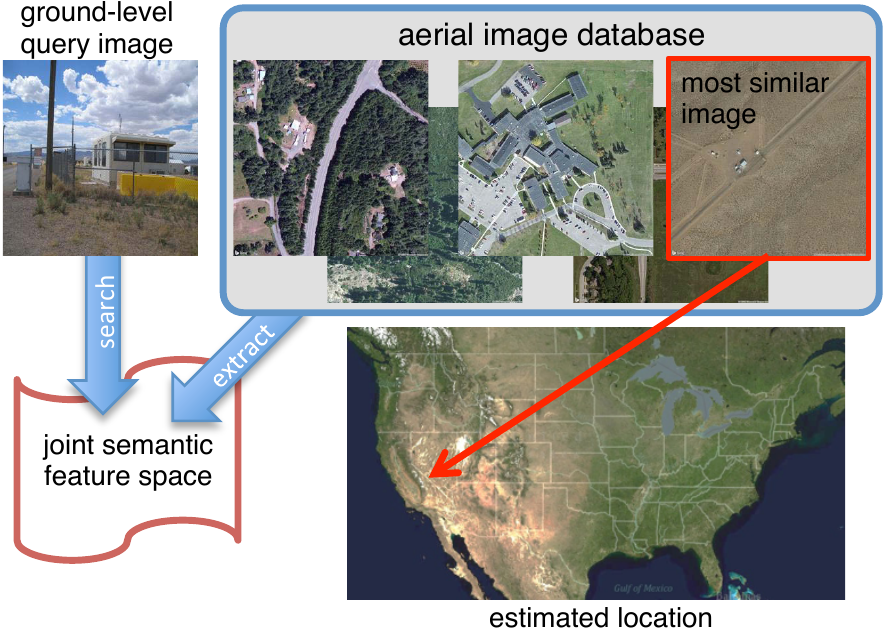}

  \caption{We learn a joint semantic feature representation for aerial and
  ground-level imagery and apply this representation to the problem of
  cross-view image geolocalization.} 

  \label{fig:cartoon}

\end{figure}

We address the problem of cross-view image
geolocalization, which aims to
localize ground-level query images by matching against a database of
aerial images (\figref{cartoon}). This contrasts with the majority of
existing image localization methods which infer location using visual
similarity between the query image and a database of other
ground-level images. The inherent limitation with these approaches is
that they fail in locations where ground-level images are not
accessible.  Even with hundreds of millions of geo-tagged ground-level
images available via photo-sharing websites and social networks, there
are still very large geographic regions with few images; most images
are captured in cities and around famous
landmarks~\cite{crandall2009mapping}. 

Cross-view image geolocalization is motivated by the observation that
the distribution of geo-tagged ground-level imagery is relatively sparse
in comparison to the abundance of high-resolution aerial imagery. The
underlying idea is to learn a mapping between ground-level and aerial
image viewpoints, such that a ground-level query image can be directly
matched against an aerial image reference database. In contrast to
previous work~\cite{lin2013cross} which used hand-engineered features,
we propose to learn feature representations using deep
convolutional neural networks (CNNs). Our methods build upon recent
success in using CNNs for ground-level image
understanding~\cite{krizhevsky2012imagenet,zhou2014places}.

We refer to our approach as cross-view training. The idea is take
advantage of existing CNNs for interpreting ground-level imagery and
use a large database of ground-level and aerial image pairs of the same
location to learn to extract semantic, geo-informative features from
aerial images.  This is a general strategy with many potential
applications but we demonstrate it in the context of cross-view
geolocalization. 

\begin{figure*}[t]

  \centering

  \includegraphics[width=\linewidth]{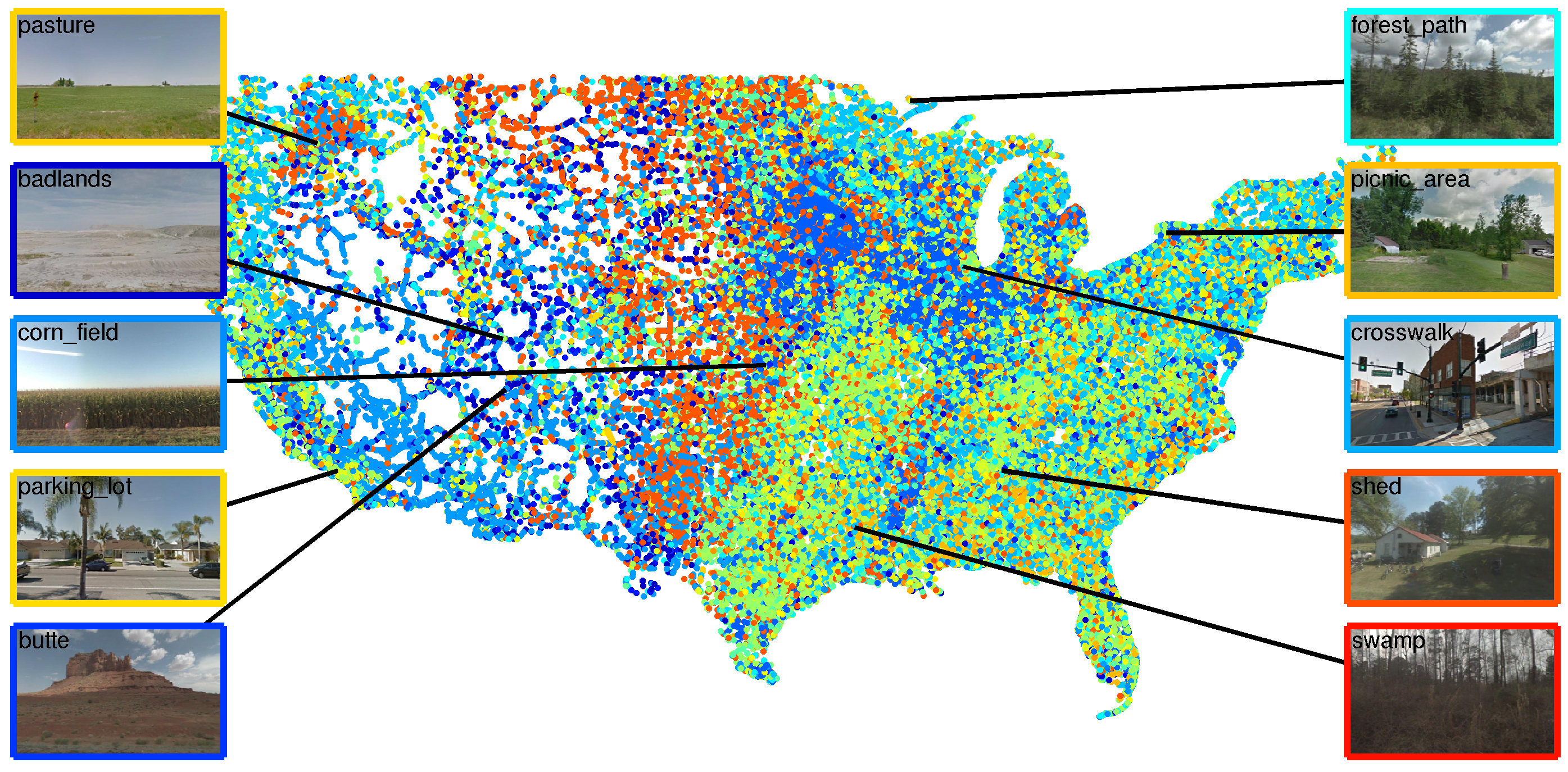}

  \caption{Existing CNNs trained on ground-level imagery provide
    high-level semantic representations which can be location
    dependent. Each point represents a geo-tagged image extracted from a
    Google Street View panorama, colored according to the predicted
    scene category  from the Places~\cite{zhou2014places} network.} 

  \label{fig:usa_streetview}

\end{figure*}

Our work makes the following main contributions: (1) an
extensive evaluation of off-the-shelf CNN network architectures and
target label spaces for the problem of cross-view localization; 
(2) cross-view training for learning a joint semantic feature space from different image sources;
(3) a massive new dataset with multi-scale aerial imagery;  
(4) state-of-the-art performance on two smaller-scale 
evaluation benchmarks for cross-view
geolocalization; and
(5) extensive qualitative evaluation, including
visualizations, which highlights the utility of cross-view training.

\section{Related Work}

Estimating the geographic location at which an image was captured based on its
appearance is a problem of great interest to the vision community. In
recent years, a plethora of methods for automatic image geolocalization
have been
introduced~\cite{baatz2012large,doersch2012what,hays2008im2gps,knopp2010avoiding,li2009landmark,zhou2014recognizing}. 
A wide variety of visual
cues have been investigated, including photometric and geometric
properties such as sun
position~\cite{cozman1995robot,lalonde2010sun,workman2014rainbow},
shadows~\cite{junejo2010gps,sandnes2011determining,wu2010geo}, and
weather~\cite{jacobs11geolocate,jacobs08geoorient,sunkavalli2008color}.

Despite this breadth, the dominant paradigm  
is to formulate the localization problem as image retrieval. 
The premise is to take advantage of the ever-increasing
number of publicly available geo-tagged images by building a large
reference dataset of ground-level images with known location. Then,
given a query image, infer its location by finding visually similar images in the dataset. 
These methods generally fall into one of two categories. The first
category of methods infer location by matching using local image
features~\cite{baatz2012large,chen2011city,crandall2009mapping,schindler2007city,snavely2006photo,torii2013visual,zamir2010accurate}.
The second category of methods match using global image
features~\cite{hays2008im2gps,jacobs07geolocate,zhou2014recognizing}.
Matching with local image descriptors is advantageous in that a more
precise location estimate is possible, but often requires additional
computational resources and fails when no visual overlap exists with
the reference dataset.  Conversely, whole image descriptors provide a
weaker prior over location but require less computation and provide a
foundation for many other image understanding tasks. 

Estimating geographic information from a single image match requires
learning geographically discriminative, location-dependent
features~\cite{doersch2012what,fang2014discovering,islam2014exploring,quercia2014aesthetic}.
The recent surge of deep learning in computer vision has shown that
convolutional neural networks can learn feature hierarchies that
perform well for a wide variety of tasks, including object
recognition~\cite{krizhevsky2012imagenet}, object
detection~\cite{girshick2014rich}, and scene
classification~\cite{zhou2014places}. Razavian et
al.~\cite{razavian2014cnn} further show that these feature hierarchies
are useful as generic descriptors. Lee et
al.~\cite{lee2014predicting} estimate geo-informative attributes from
an image using convolutional neural network classifiers.

Only recently has aerial imagery been discovered as a valuable resource 
for ground-level image
understanding~\cite{bansal2011geo,luo2008event}.  Shan et
al.~\cite{shan2014accurate} geo-register ground-level multi-view
stereo models using ground-to-aerial image matching. Viswanathan et
al.~\cite{viswanathan2014vision} evaluate a number of hand-engineered
feature descriptors for the task of ground-to-aerial image matching in
robot self-localization. The cross-view image geolocalization problem
was introduced by Lin et al.~\cite{lin2013cross}. Workman et
al.~\cite{workman2015geocnn} show that features extracted from
convolutional neural networks are useful for problems in geospatial
image analysis. Most akin to our work, Lin et
al.~\cite{lin2015learning} apply a siamese CNN architecture for
learning a joint feature representation between ground-level images
and $45^\circ$ oblique aerial imagery. Our approach is more general;
we operate on orthorectified aerial imagery, do not require scale and
depth metadata for each query, and our joint feature representation is
semantic.

\section{Cross-View Training for Aerial Image Feature Extraction}
\label{sec:formulation}

We propose a cross-view training strategy that uses deep convolutional
neural networks to extract features from aerial imagery. The
key idea is to use pre-existing CNNs for extracting ground-level image
features and then learn to predict these features from aerial images
of the same location. This is a general approach that could be useful
in a wide variety of domains. It is conceptually similar to domain
adaptation~\cite{daume2006domain}, where the source domain is the
ground-level view and the target domain is aerial imagery. The end
result of cross-view training is a CNN that is able to extract
semantically meaningful features from aerial images without manually
specifying semantic labels. 

\subsection{Cross-View Feature Representations}

We assume the existence of two functions: $f_a(l;\Theta_a)$, which
extracts features from the aerial imagery centered at location, $l$,
and $f_g(I;\Theta_g)$, which extracts features from a ground-level
image. Here, $\Theta_g$ and $\Theta_a$ are the parameters for feature
extraction. We propose to use deep feed-forward convolutional neural
networks as the feature extraction functions, $f_a$ and $f_g$. In this
framework, the parameters of these functions, $\Theta_a$ and
$\Theta_g$, include both the network architecture and the weights.

Our main insight is that we can take advantage of the significant
progress that has been made applying CNNs to ground-level image
understanding in the past several years by \emph{transferring} feature
representations to aerial images. This is possible if the location of
the ground-level imagery is known. For example, in \figref{usa_streetview},
we show the estimated label from the Places~\cite{zhou2014places}
network, trained for the task of scene classification, on a set of
images extracted from Google Street View panoramas captured across the
United States. The predicted label is clearly location dependent.
For the purposes of learning a useful aerial image feature function,
what matters is that the ground-level features are geo-informative,
not necessarily that the ground-level detector is perfect.

We compare alternative choices for ground-level feature extraction in
\secref{evaluation} for the problem of cross-view image
geolocalization. In the remainder of this section, we describe our
cross-view training approach to adapt a network trained for
ground-level feature extraction to aerial imagery.

\subsection{Cross-View Training a Single-Scale Model}

Given a semantically meaningful feature representation for ground
imagery, we propose to 
 extract features from aerial imagery, which we
refer to as cross-view training. Given a set of
ground-level training images, $\{I_i\}$, with known location,
$\{l_i\}$ and known ground-level feature extractor
parameters, $\Theta_g$, we seek a set of parameters, $\Theta_a$, that
minimize the following objective function:
\begin{equation}
  J(\Theta_a) = \sum_i \|f_a(l_i;\Theta_a) - f_g(I_i;\Theta_g)\|_2.
  \label{eq:objective}
\end{equation}
Intuitively, the objective is to learn to extract features from the
aerial imagery that match those from a corresponding ground-level
image. 

\subsection{Cross-View Training a Multi-Scale Model}

The view frustum of ground-level imagery can vary dramatically from
image to image. It is possible that the nearest object in the scene is
hundreds of meters away or that the furthest object is tens of meters.
This introduces ambiguity when matching the location observed by a
ground-level image to the known geolocation of the aerial imagery. To
address this issue, we extend our aerial image feature function,
$f_a$, to support extracting features at multiple spatial scales. Rather
than mapping a single ground-level image to a single aerial image, the
multi-scale approach allows for a ground-level image to be matched
to aerial images at multiple scales. 
In support of multi-scale, cross-view training, we introduce a large dataset of
ground-level and aerial image pairs. 

\subsection{A Large Cross-View Training Dataset}

\begin{figure}

  \centering

  \begin{subfigure}[b]{.49\linewidth}
    \centering
    \includegraphics[width=\linewidth]{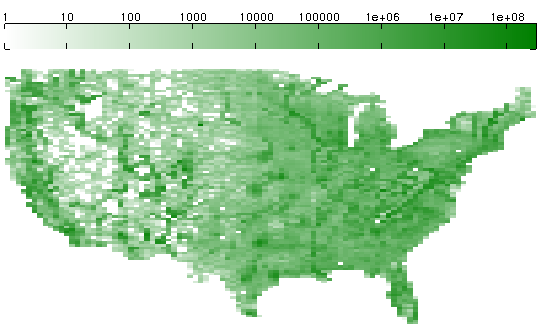}
    \caption{Google Street View}
  \end{subfigure}
  \begin{subfigure}[b]{.49\linewidth}
    \centering
    \includegraphics[width=\linewidth]{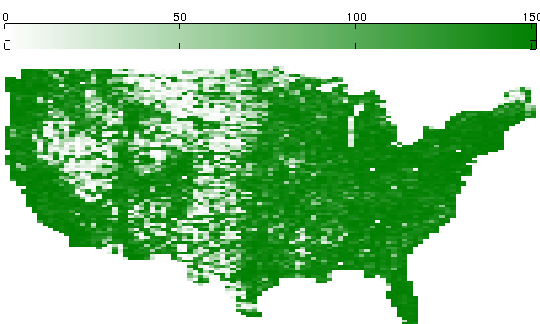}
    \caption{Flickr}
  \end{subfigure}
  
  \caption{The distribution of ground-level images in the CVUSA
  dataset.} 

  \label{fig:usa_dataset}

\end{figure}

\begin{figure*}[t]
  \centering
  
  \includegraphics[width=.12\textwidth]{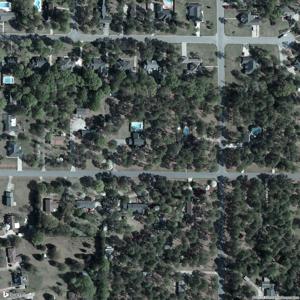}%
  \includegraphics[width=.12\textwidth]{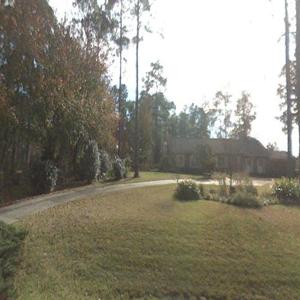}%
  \hfill
  \includegraphics[width=.12\textwidth]{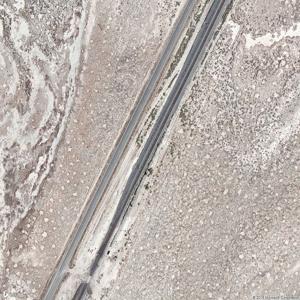}%
  \includegraphics[width=.12\textwidth]{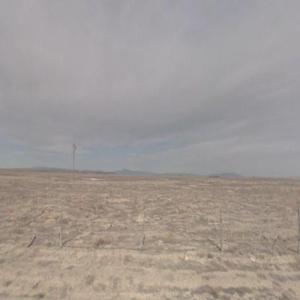}%
  \hfill
  \includegraphics[width=.12\textwidth]{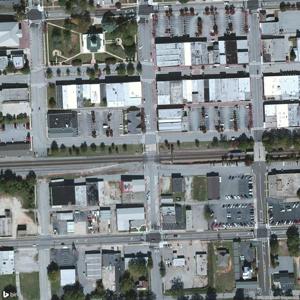}%
  \includegraphics[width=.12\textwidth]{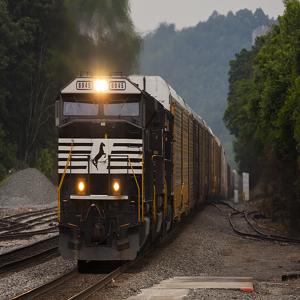}%
  \hfill
  \includegraphics[width=.12\textwidth]{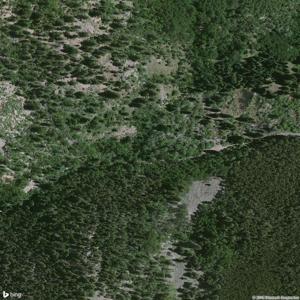}%
  \includegraphics[width=.12\textwidth]{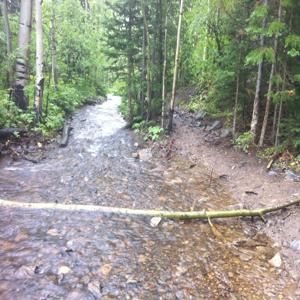}%
  
  \caption{Example matched ground-level and aerial images from the
  CVUSA dataset.}%
  \label{fig:dataset}%
\end{figure*}

Previous cross-view datasets have been limited in spatial scale and
number of training images. The largest
dataset~\cite{workman2015geocnn} contains $174\,217$ training image pairs
sampled from a $200km \times 200km$ area around San Francisco.
Features learned using such a dataset are unlikely to be as effective
when applied to another location. In an effort to broaden the
applicability of the learned feature extractor, we constructed a
massive dataset of pairs of ground-level and aerial images from across the
United States, called the Cross-View USA (CVUSA) dataset.

Geo-tagged, ground-level images were collected from both Google Street View and Flickr.
For Google Street View, we randomly sampled from locations 
within the continental United States. At each location, we obtained the corresponding panoramic image
and extracted two perspective images from viewpoints separated by 180$^\circ$ along the roadway. 
For Flickr, we divided the area of the United States into
a $100 \times 100$ grid and downloaded up to 150 images from each grid
cell (from 2012 onwards, sorted by the Flickr ``interesting'' score). As Flickr images are 
overrepresented in urban areas, this binning step ensures a more even 
sampling distribution. From this set, we automatically filtered out
images of indoor scenes using the {\em Places}~\cite{zhou2014places}
scene classification network by retaining images that match to one of
the outdoor scene categories. 

This process resulted in $1\,036\,804$ Street View images and $551\,851$
Flickr images.  \figref{usa_dataset} visualizes the relative density
of each set of images. For each ground-level image, we
downloaded an $800 \times 800$ aerial image centered at that location
from Bing Maps, at multiple spatial scales (zoom levels 14, 16 and 18).
After accounting for overlap, this results in $879\,318$ unique aerial
image locations and a total of $1\,588\,655$ million geo-tagged, image matched pairs.
\figref{dataset} shows several example matched ground-level and aerial images
from our dataset.

\section{Application to Cross-View Localization}
\label{sec:evaluation}

We focus on the problem of cross-view image
geolocalization~\cite{lin2013cross} in which the goal is to use a
database of aerial images, with known location, to estimate the
geographic location of a ground-level query image in that region. This
is a challenging problem because of the dramatic appearance
differences between ground-level and aerial viewpoints. 

\subsection{Evaluation Datasets}

We evaluate our proposed cross-view training approach on two existing benchmark
datasets. The first dataset, Charleston, was introduced by Lin et
al.~\cite{lin2013cross} and contains imagery from a $40km \times 40km$
region around Charleston, South Carolina. In total, there are $6\,756$
ground-level images collected from Panoramio, each with an associated
aerial image and land-cover attribute map centered at its location.
The aerial image reference database contains $182\,988$ images.  The
second benchmark dataset, San Francisco, is introduced by Workman et
al.~\cite{workman2015geocnn} and contains imagery from a $200km
\times 200km$ region around San Francisco, California. Ground-level imagery
consists of $74\,217$ images from Flickr and $100\,000$ Street View cutouts.
Similar to Charleston, each ground-level image is accompanied by a
corresponding aerial image centered at the ground-level image location. The
aerial image reference database contains $278\,561$ images. Each dataset
identifies a set of ``hard to localize'' ground-level images, with no nearby
ground-level reference imagery, to be used for evaluation. 

\subsection{Localization Method and Performance Metric}
The process for localizing a ground-level query image,
$\hat{I}$, is straightforward. We directly compare the ground-level feature,
$f_g(\hat{I};\Theta_g)$, for the query image against a reference
aerial image feature, $f_a(l;\Theta_a)$, at location $l$, using
Euclidean distance $\|f_a(l;\Theta_a) - f_g(\hat{I};\Theta_g)\|_2$.
If a single pinpoint match is needed, we return the geolocation of
the image
that is the nearest
neighbor of the ground-level image in feature space; otherwise we return a list of candidate regions sorted by
distance in feature space.
As described by Lin et
al.~\cite{lin2013cross}, the performance metric for this problem
is the rank of the ground truth location in the sorted list
of localization scores, for a set of
aerial image reference locations. We represent the
localization results using a cumulative
graph of the percentage of correctly
localized images as a function of the percentage of candidates searched.

\subsection{Localization using Off-The-Shelf CNN Features}

As a baseline to our cross-view training approach, we evaluated the
localization performance of ``off-the-shelf'' CNN features on
Charleston. We extracted features from both the aerial and
ground-level query image using a variety of network architectures
trained for different target label spaces.  The network architectures
used included GoogleNet~\cite{szegedy2015going},
AlexNet~\cite{krizhevsky2012imagenet}, NIN~\cite{lin2014network}, and
VGG 19~\cite{chatfield2014return}. Training databases included
Places~\cite{zhou2014places}, ImageNet~\cite{russakovsky2015imagenet},
Hybrid~\cite{zhou2014places}, Oxford
Flowers~\cite{nilsback2008automated}, and Flickr
Style~\cite{karayev2014recognizing}.  We evaluated multiple such
configurations, all publicly available as Caffe~\cite{jia2014caffe}
model files. 

\begin{figure}

  \centering

  \includegraphics[width=\linewidth]{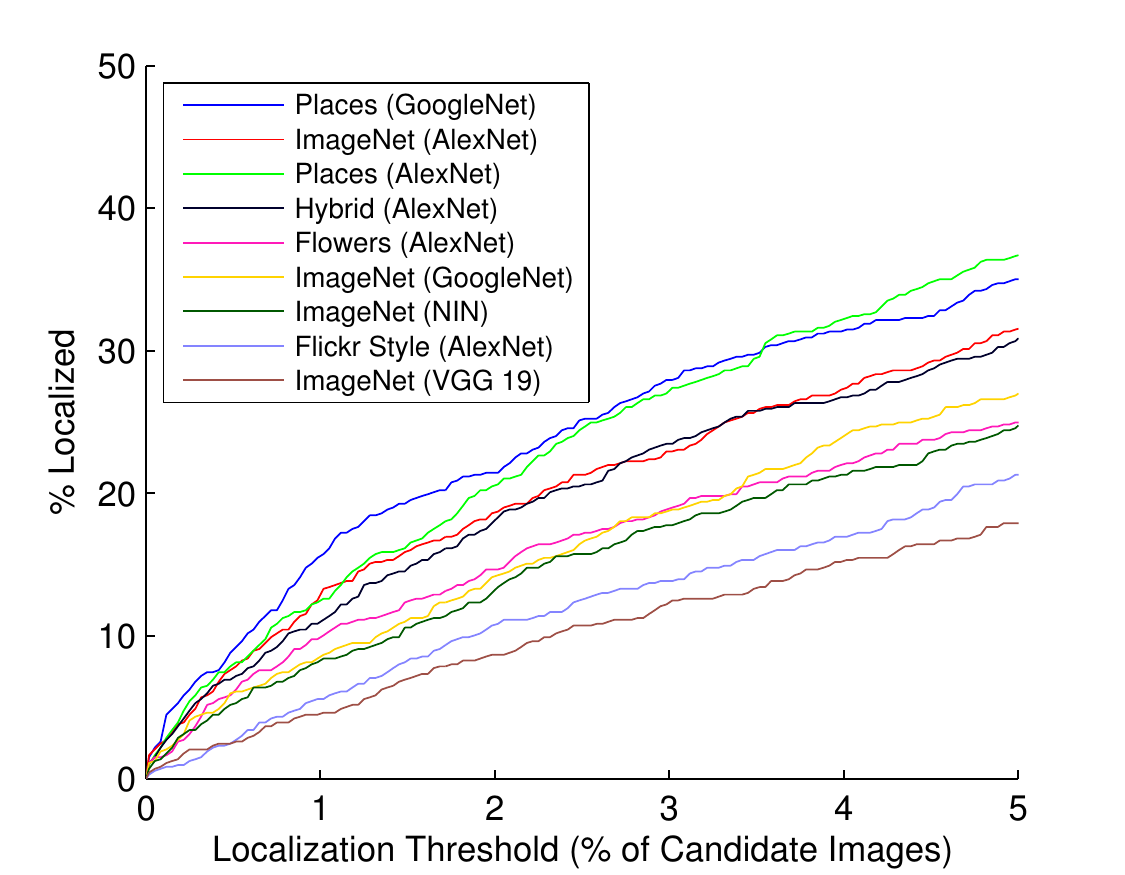}

  \caption{Comparison of several off-the-shelf CNN features in terms
  of localization accuracy on the Charleston dataset.} 

  \label{fig:baseline}

\end{figure}

Our findings from this experiment are visualized in \figref{baseline}.
The top two performing configurations in terms of top 5\% accuracy
are trained for the task of
scene classification on the Places~\cite{zhou2014places} database,
which contains 
over two million images labeled from 205 different categories. These
two networks vastly outperform the next best network, which was
trained on ImageNet for the task of object recognition.  These results
are interesting, but unsurprising, as scenes are more likely to be
visible from aerial imagery.  For the rest of the experiments, we
apply cross-view training to learn an aerial image feature extractor
for Places features using the {\em AlexNet}
architecture~\cite{zhou2014places}, which we refer to as {\em Places}. 

\subsection{Localization using Cross-View Features}

The {\em AlexNet} architecture~\cite{krizhevsky2012imagenet} consists of five 
convolutional layers (interspersed with dropout, pooling, and local
response normalization layers) and three fully-connected layers (called
`fc6', `fc7', and, the output layer, `fc8').  The only difference
with {\em Places} is the dimensionality of the output layer (205
versus 1000 possible categorical labels). 

\begin{figure}

  \centering

  \begin{subfigure}[b]{.9\linewidth}
    \centering
    \includegraphics[width=\linewidth]{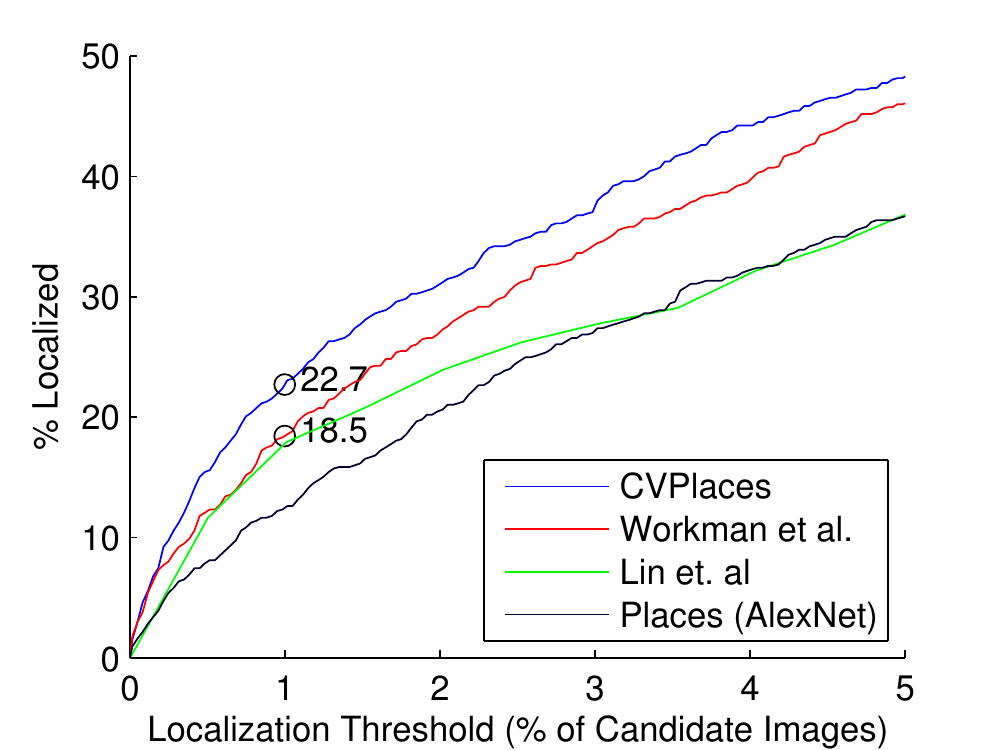}
    \caption{Charleston}
  \end{subfigure}
  \begin{subfigure}[b]{.9\linewidth}
    \centering
    \includegraphics[width=\linewidth]{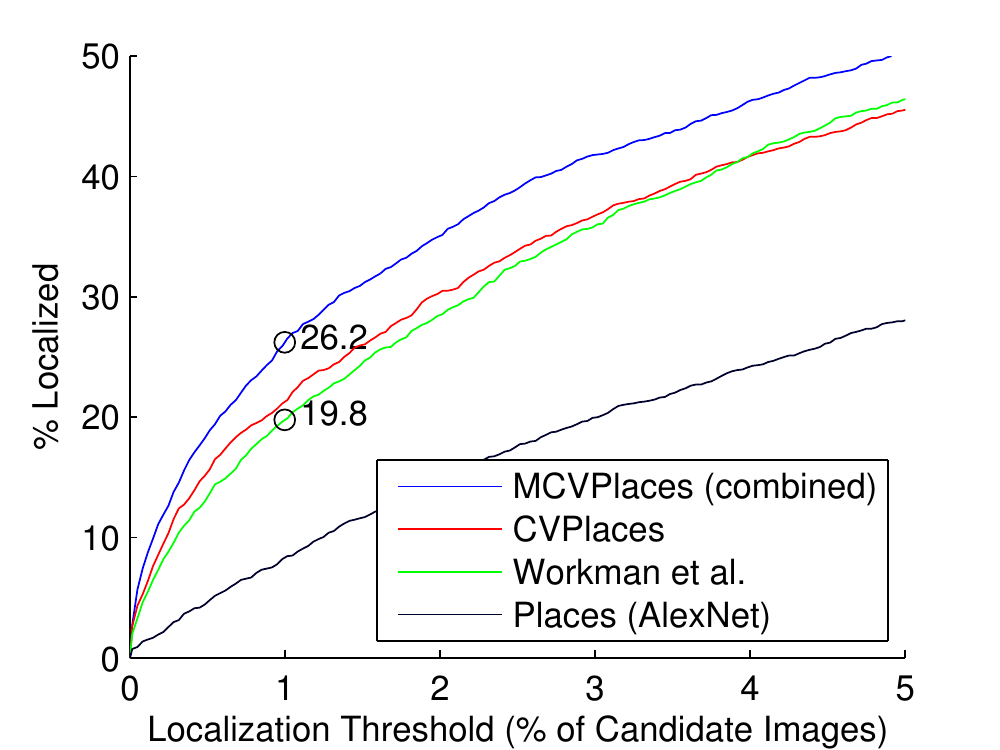}
    \caption{San Francisco}
  \end{subfigure}
  
  \caption{Accuracy of localization as a function of retrieved
    candidate locations on two benchmark datasets.} 

  \label{fig:cdf}

\end{figure}

Given the architecture and weights, $\Theta_g$, of {\em Places}, we apply
the cross-view training approach described in \secref{formulation} to
train a model to predict the `fc8' features. In practice, we fix the
network architecture and optimize the weights. For training, we
use pairs of ground-level images and the highest-resolution aerial
images in our CVUSA dataset (zoom level 18). We refer to this model as
{\em CVPlaces}. \figref{cdf} shows the improvement in localization of
our single-scale model, with and without cross-view
training, on Charleston and San Francisco. 

Initial experiments showed that initializing the solver with
$\Theta_a^0 = \Theta_g$ worked well, therefore we use that strategy
throughout.  We reserve 1000 matched pairs of images from each
benchmarks training set as a validation set for model selection. 
Our models are implemented using the Caffe toolbox~\cite{jia2014caffe}
and trained using stochastic gradient descent with a Euclidean loss
for parameter fitting to reflect~\eqref{eq:objective}. The full model
file, solver definition, and learned network weights are available
online.\punctfootnote{\url{http://cs.uky.edu/~scott/}}

\subsection{Evaluating Multi-Scale Cross-View Training}

Our multi-scale model architecture consists of three single-scale {\em
CVPlaces} networks with untied weights, each taking as input a
different spatial resolution of aerial imagery. The top feature layer
from each individual network is concatenated and used as input to a
final fully-connected layer with a 205 dimensional output. The
resulting model has approximately 180 million parameters.  For
training, we initialize each of the sub-networks with the weights for
our best single-scale network and randomly initialize the output
layer. We refer to our multi-scale model as {\em MCVPlaces}.

To evaluate {\em MCVPlaces}, we augmented San Francisco with
additional multi-scale aerial imagery (zoom levels 16 and 14).
\figref{cdf} shows a comparison of our multi-scale approach versus our
single-scale approach and a recent method on San Francisco. The
features learned via multi-scale cross-view training significantly
out-perform all others. In terms of top 1\% accuracy, we improve the
state-of-the art by 6.4\%, a percentage change of 32.32\%.

\begin{figure*}

\centering

\bgroup
\setlength\tabcolsep{0cm}
\def\arraystretch{.2}
\begin{tabular}{ccccccc}
  wheat field &
  arch &
  airport & 
  apt.\ building & 
  cemetery & 
  lighthouse & 
  badlands
\\
\hline
\multicolumn{7}{c}{ground-level images on the {\em Places} network~\cite{zhou2014places}}
\\
\includegraphics[width=.14\textwidth]{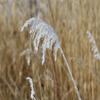}%
&
\includegraphics[width=.14\textwidth]{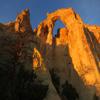}%
&
\includegraphics[width=.14\textwidth]{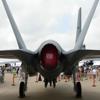}%
&
\includegraphics[width=.14\textwidth]{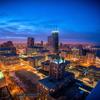}%
&
\includegraphics[width=.14\textwidth]{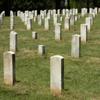}%
&
\includegraphics[width=.14\textwidth]{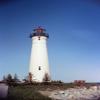}%
&
\includegraphics[width=.14\textwidth]{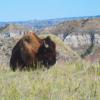}%
\\
\multicolumn{7}{c}{aerial images on the {\em Places
network}~\cite{zhou2014places}}
\\
\includegraphics[width=.14\textwidth]{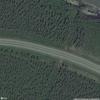}%
&
\includegraphics[width=.14\textwidth]{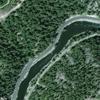}%
&
\includegraphics[width=.14\textwidth]{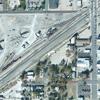}%
&
\includegraphics[width=.14\textwidth]{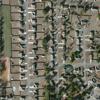}%
&
\includegraphics[width=.14\textwidth]{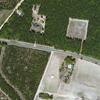}%
&
\includegraphics[width=.14\textwidth]{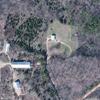}%
&
\includegraphics[width=.14\textwidth]{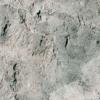}%
\\
\multicolumn{7}{c}{aerial images on our {\em CVPlaces} network}
\\
\includegraphics[width=.14\textwidth]{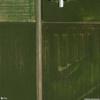}%
&
\includegraphics[width=.14\textwidth]{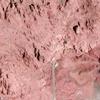}%
&
\includegraphics[width=.14\textwidth]{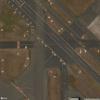}%
&
\includegraphics[width=.14\textwidth]{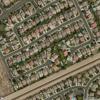}%
&
\includegraphics[width=.14\textwidth]{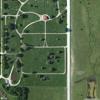}%
&
\includegraphics[width=.14\textwidth]{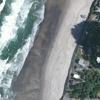}%
&
\includegraphics[width=.14\textwidth]{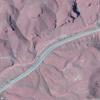}%

\end{tabular}
\egroup

\caption{Images that result in high activations for particular scene
  categories. (top) The high-activation ground-level images are
  exemplars for the corresponding semantic class.  (middle)
  The high-activation aerial images for the network trained on
  ground-level images are, not surprisingly, less semantically
  correct.  For example, in the ``arch'' category the image may look
  like an arch, but is not a location you are likely to see an arch
  from the ground.  (bottom) After fine-tuning for the aerial domain, the
  high-activation images are a better match to the respective categories.}

\label{fig:extreme}

\end{figure*}

\section{Discussion}
\label{sec:discussion}

The evaluation suggests that the cross-view training procedure learns
features that are effective for localization. In the
remainder of this section, we explore this representation in
more depth.

\subsection{Understanding Network Activations}

To understand what the network is learning, we analyze the node-level activations for a large set of images on the
{\em Places} network and our {\em CVPlaces}
network. We randomly sampled $20\,000$ pairs of ground-level/aerial
images from CVUSA and recorded the activations for each.
\figref{extreme} shows a set of images that resulted in the maximum
activation for particular `fc8' nodes of each network. We selected the
`fc8' nodes because they are the last layer before the {\em softmax}
output and are therefore semantically meaningful. The ground-level
images that result in high activations on the {\em Places}
network are
good exemplars of their corresponding category. However, using the same network, 
high-activation aerial images are often
semantically incorrect. For example the ``wheat field'' image is
actually a forest and the ``airport'' image is a highway. When passed
through our {\em CVPlaces} network, the high-activation images are much
more semantically plausible. These results highlight that the
cross-view training process is learning to recognize locations in
aerial images where particular scene categories are likely to be observed 
from a ground-level viewpoint. 

\subsection{Geospatial Visualization of Aerial Image Features}

\begin{figure}

  \centering

  \includegraphics[width=.495\linewidth]{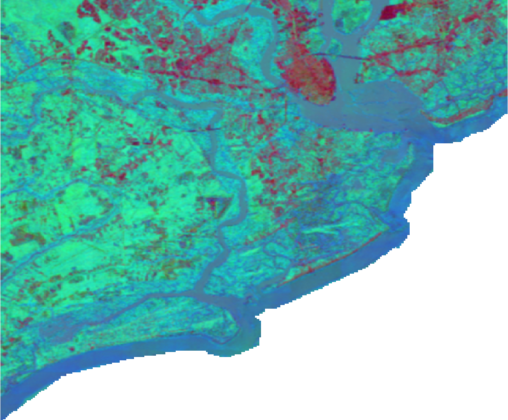} 
  \hfill
  \includegraphics[width=.495\linewidth]{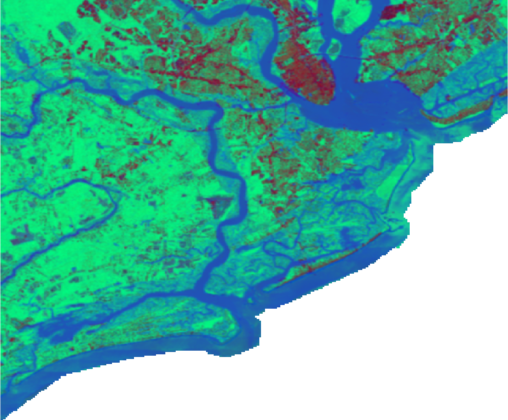} 

  \caption{
    (left) A false-color image generated by applying the {\em Places}
    network to aerial imagery.  In both images the colors are
    semantically meaningful
    (red=urban, green=rural, blue=water-related).  (right) The same as
    (left) but with our {\em CVPlaces} network (trained on the entire
    USA dataset, with no Charleston-specific fine tuning).}

\label{fig:falsecolor}

\end{figure}

We visualize the geospatial distribution of high-level features
extracted from the high-resolution aerial reference imagery from the
Charleston dataset~\cite{lin2013cross}.  The result is a
coarse-resolution false-color image that summarizes the semantic
information extracted by a particular CNN from the aerial images. To support this, we
computed the `fc8' features from two networks, {\em Places} and our
{\em CVPlaces}.  For visualization purposes, we choose three
high-level categories (urban, rural, and water-related) and
assign a set of representative scene categories to each. 
The false-color image is generated as follows: for the red channel, we compute the
average activation for the set of categories defined as urban on the
aerial imagery under each pixel. The same procedure is applied for rural (green) and
water-related (blue). We then linearly scale the averaged activations
to the range $[0,1]$. The result is a false-color aerial image
(\figref{falsecolor}) with semantically meaningful colors. For
example, a bright red pixel identifies an urban area and a purple
pixel is an urban area near the water, etc. Our {\em CVPlaces} network
results in a clearer distinction between regions, highlighting
the urban core of Charleston and distinguishing water regions from
rural. This demonstrates that the cross-view training procedure enables
the {\em CVPlaces} network to extract semantically meaningful features
from aerial imagery. This is especially interesting because the
network was trained using the entire CVUSA dataset and was not
fine-tuned specifically for the Charleston area.

\subsection{Localization at Dramatically Different Spatial Scales}

\begin{figure}

  \centering

  \includegraphics[height=.8in]{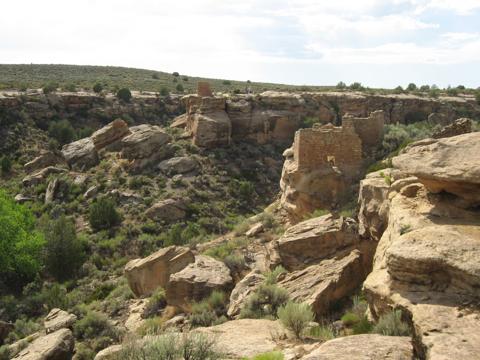}
  \includegraphics[height=.8in]{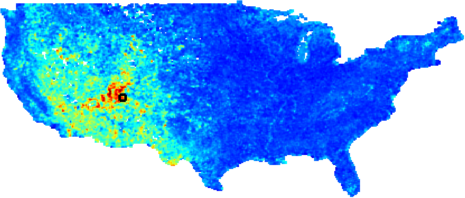}

  \includegraphics[height=.8in]{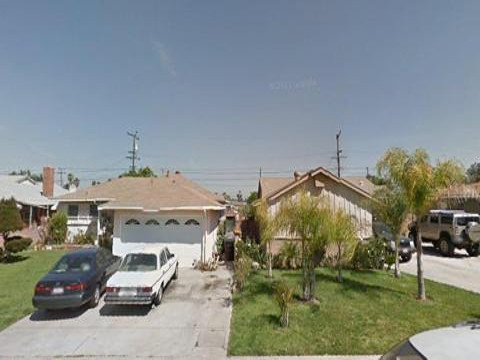}
  \includegraphics[height=.8in]{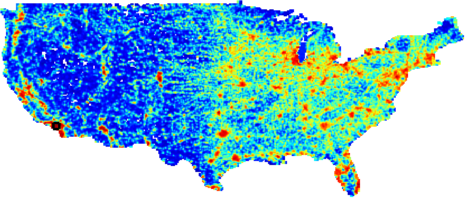}

  \includegraphics[height=.8in]{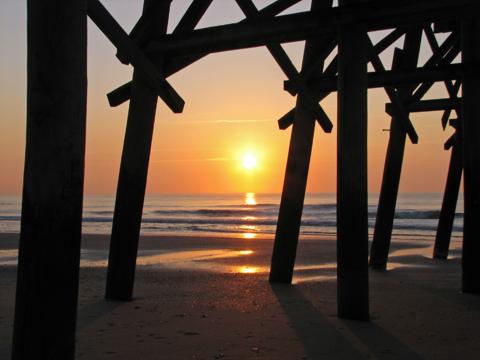}
  \includegraphics[height=.8in]{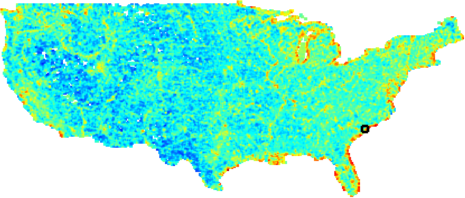}

  \caption{Localization examples at a continental scale. (left) A
    ground-level query image.  (right) A heatmap of the distance
    between the {\em Places} `fc8' feature of the query image and the
    corresponding {\em CVPlaces} feature of an aerial image at that
    location (red: more likely location, blue: less likely location).
    The black circle marks the true location of the camera.} 

  \label{fig:usa}

\end{figure}

The quantitative evaluation shows that by using our {\em CVPlaces}
network, we obtain state-of-the-art localization performance at the
scale of a major metropolitan area (approx. $100km$ across). In this
section, we explore whether {\em CVPlaces} might work at larger and
smaller spatial scales. We begin at the continental scale: given a
ground-level query image from CVUSA, we compute the feature distance
between the {\em Places} `fc8' feature vector of the query image and
{\em CVPlaces} `fc8' feature vector of all aerial images in the
dataset. \figref{usa} shows qualitative results as a heatmap that
represents the distance between the query and corresponding aerial
image. The black dot represents the ground truth location of the query
images. In the first example, our method clearly identifies the image
as having been captured in the desert southwest. The second example,
of a suburban neighborhood, results in a heatmap that highlights urban
areas. The third example identifies the query image as having been
captured on a coast.

\begin{figure*}

  \centering

  \begin{tabular}{ccccc}
    \includegraphics[width=.18\textwidth]{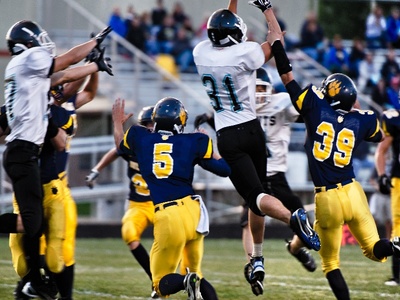} &
    \includegraphics[width=.18\textwidth]{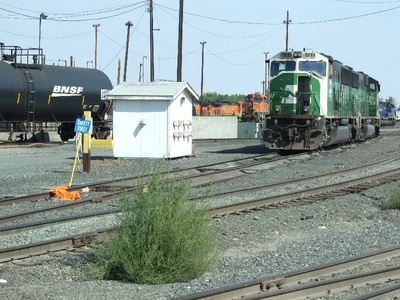} &
    \includegraphics[width=.18\textwidth]{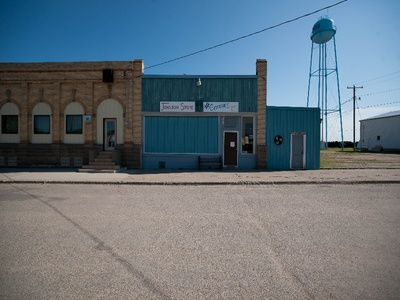} &
    \includegraphics[width=.18\textwidth]{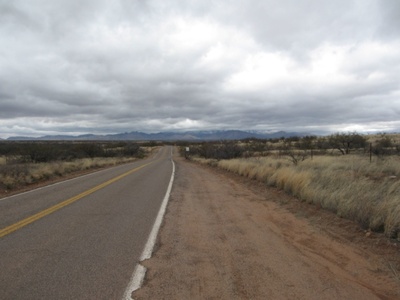} & 
    \includegraphics[width=.18\textwidth]{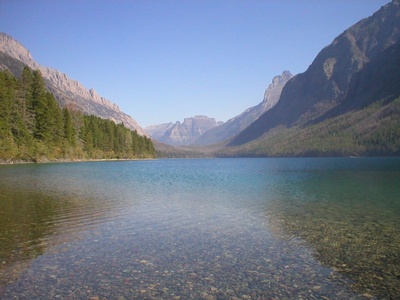} 
    \\
    \includegraphics[width=.18\textwidth]{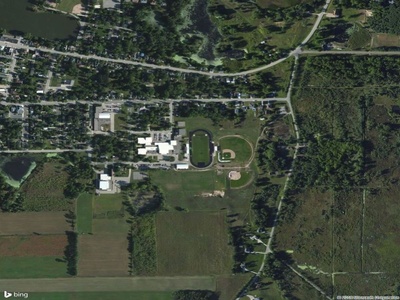} &
    \includegraphics[width=.18\textwidth]{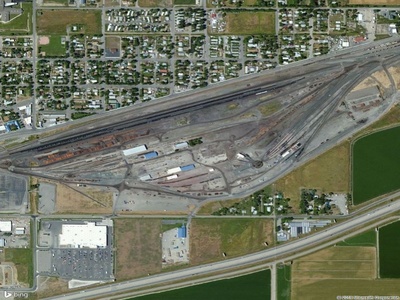} &
    \includegraphics[width=.18\textwidth]{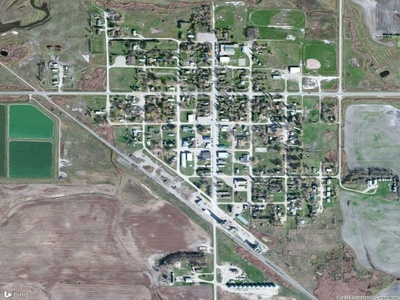} &
    \includegraphics[width=.18\textwidth]{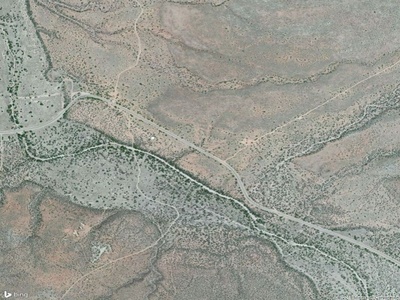} &
    \includegraphics[width=.18\textwidth]{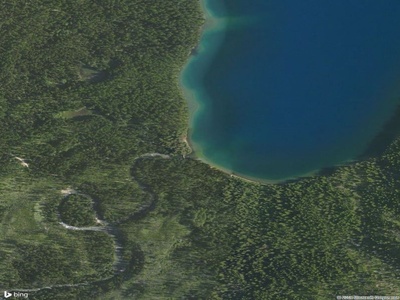} 
    \\
    \includegraphics[width=.18\textwidth]{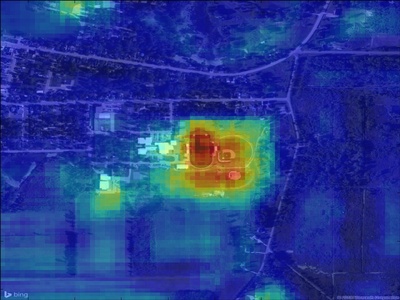} &
    \includegraphics[width=.18\textwidth]{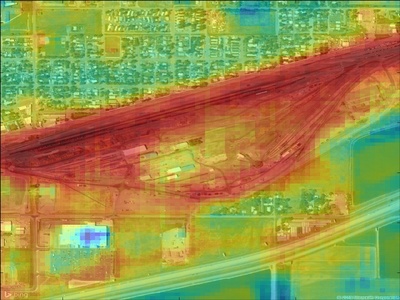} &
    \includegraphics[width=.18\textwidth]{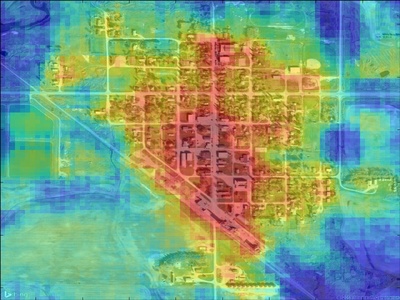} &
    \includegraphics[width=.18\textwidth]{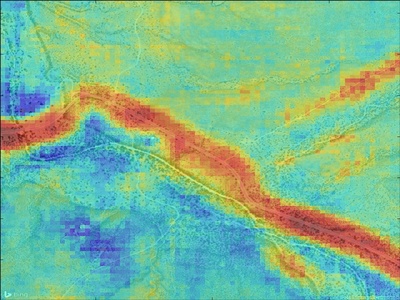} &
    \includegraphics[width=.18\textwidth]{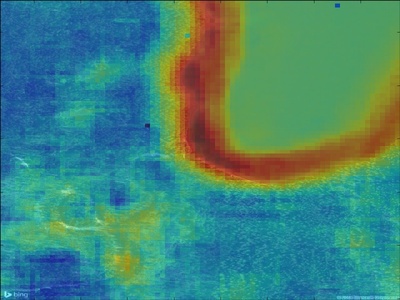} 
  \end{tabular}

  \caption{Examples of localization at finer spatial scales. (top) The
    ground-level query image. (middle) An aerial image centered at the
    ground location. (bottom) An overlay showing the distance between
    the ground-level image feature and the aerial image features at each
    location, computed using a sliding window approach (red: more
    likely, blue: less likely).}

\label{fig:fine}

\end{figure*}

We also explore whether the proposed method can be used for 
localization at a much smaller scale.  \figref{fine} shows examples
where the method is able to distinguish between locations a few
decameters apart. To accomplish this, we implemented a system that
takes as input a query image and an initial location estimate. It
samples a grid of nearby geographic locations and computes the
distance between the {\em Places} `fc8' feature vector of the query
image and the corresponding {\em CVPlaces} feature of the sub-window
of the aerial imagery. Note that sampling on the grid could be
accelerated by computing it convolutionally on the GPU.  These results
show that in some cases, such as the American football example, it can
identify a football stadium given an image of players. In the other examples, the heatmaps
reflect the inherent uncertainty of localization. The lake-shore
example is particularly interesting because even though the shore is
not visible, the heatmap correctly reflects that the photographer is
less likely to be standing in the middle of the lake than on its
shore.

\section{Conclusion}

We proposed a cross-view training approach, in which we learn to
predict features extracted from ground-level imagery from aerial
imagery of the same location. We introduced a massive dataset of such
pairs and proposed single and multi-scale networks
for extracting aerial image features, obtaining  state-of-the-art
results for cross-view localization on two benchmark datasets.

Our focus was learning the optimal parameters, $\Theta_a$, for
extracting features from aerial imagery. We tried fixing the aerial
parameters, $\Theta_a$, using pre-existing networks, and optimizing
over $\Theta_g$, but the performance was poor. We also attempted
jointly optimizing over $\Theta_a$ and $\Theta_g$ but the results did
not improve over exclusively optimizing for $\Theta_a$. We suspect
both of these results are because existing ground-level image
feature extractors are better suited for cross-view localization than
aerial image feature extractors. However, finding better
initial values for $\Theta_a$ is an interesting area for future work. 

When the ground-level query image was captured in a location that is
distinctive from above, such as an outdoor football stadium or an
intersection with a unique pattern of intersecting roads, it is
possible to obtain a precise estimate of the geographic location using
the cross-view localization approach. However, many locations are not
so distinctive. Therefore, it is useful to consider the proposed
approach as a pre-processing step to a more expensive matching
process. Such a matching process might be purely computational, as
with sparse keypoint matching, or may involve manual human search. 

\ificcvfinal
\section*{Acknowledgements} Supported by the Intelligence Advanced
Research Projects Activity (IARPA) via Air Force Research Laboratory,
contract FA8650-12-C-7212.  The U.S. Government is authorized to
reproduce and distribute reprints for Governmental purposes
notwithstanding any copyright annotation thereon. Disclaimer: The
views and conclusions contained herein are those of the authors and
should not be interpreted as necessarily representing the official
policies or endorsements, either expressed or implied, of IARPA, AFRL,
or the U.S. Government.
\fi

{
\small
\bibliographystyle{ieee}
\bibliography{biblio}
}

\end{document}